# USING HOLOGRAPHICALLY COMPRESSED EMBEDDINGS IN QUESTION ANSWERING

Salvador E. Barbosa

Department of Computer Science, Middle Tennessee State University,
Murfreesboro, TN, USA

*ABSTRACT*

*Word vector representations are central to deep learning natural language processing models. Many forms of these vectors, known as embeddings, exist, including word2vec and GloVe. Embeddings are trained on large corpora and learn the word's usage in context, capturing the semantic relationship between words. However, the semantics from such training are at the level of distinct words (known as word types), and can be ambiguous when, for example, a word type can be either a noun or a verb. In question answering, parts-of-speech and named entity types are important, but encoding these attributes in neural models expands the size of the input. This research employs holographic compression of pre-trained embeddings, to represent a token, its part-of-speech, and named entity type, in the same dimension as representing only the token. The implementation, in a modified question answering recurrent deep learning network, shows that semantic relationships are preserved, and yields strong performance.*

## 1. INTRODUCTION

Question answering (QA) is the natural language processing (NLP) task of selecting a span of text from one or more context documents, as the answer to a specific question. The context documents, referred to as *passages* in this paper, may be anything from a single sentence to a collection of multiple documents. Early QA systems were symbolic and rule-based, but recently, deep neural networks have become the standard for such systems, yielding excellent performance while avoiding the brittleness of rule-based systems. The distributed representations learned in these networks have been significantly enhanced by vector representations of word tokens, known as embeddings.

Embeddings are dense, continuous vectors, learned from text corpora, which capture the semantic relatedness of individual words from the context of surrounding words. It should be noted that the relationship extracted here is not necessarily a synonymic one, but is rather one of being in the company of similar words. Thus, the words *love* and *hate* are usually identified as highly related embeddings, even though they are antonyms of each other, because they tend to appear is similar contexts.

Commonly used embeddings in NLP include *word2vec* [1] and *GloVe* [2]. These embeddings are pre-trained on large corpora (such as the Wikipedia text), and represent tokens as real-valued vectors with 50 to 300 dimensions. Deep neural networks utilize vectors of individual tokens to represent words, phrases, sentences, and higher aggregations of text, depending on the application and underlying network topology. Embeddings are mostly computed at the level of distinct words (known as word types). As such, there is a single vector for a word, regardless of whether it may have more than one part-of-speech in its various uses in the training corpus. Because of their importance to the task, many QA systems use the part-of-speech tag and named entity type (Person, Organization, Money, Time, etc.) of individual words as separate features, in extracting an answer. Doing so, however, expands the dimension of the input for each token, potentially leading to the curse of dimensionality. This research eliminates the expansion in the size of the input by using holographic compression on the vectors. Employing the proposed approach, the part-of-speech tag and named entity type of each token is encoded without increasing the size of the input, while providing the benefits generally associated with dimensionality reduction: a decrease in overfitting and a reduction in processing time.

Holographic reduced representations (HRR) were devised by Tony Plate as a means for combining two or more vectors of a given dimension, into a single vector having the same dimension as only one of those components [3]. Thus, unlike the concatenation of two vectors of dimension $n$ which results in a vector of dimension $2n$, HRR uses circular convolution to combine vectors, without increasing the dimensionality of the resultant vector.

Recurrent neural networks (RNN) are commonly used to learn from sequences of data, such as the words that form a sentence or larger collection of words [4]. Rather than feeding the entire sequence as a single input (concatenated or superimposed in some way), individual elements of the sequence are processed one at a time. The RNN is constructed so that at each time step, it considers the current input element, along with a combined or accumulated representation of the elements that preceded it, to update its learning. The long-short-term-memory (LSTM) architecture has become ubiquitous in question answering applications in recent times [5] – [7].

DrQA is a question answering framework developed by Facebook Research, to answer open domain questions from sources such as Wikipedia [8]. It is made up of two major components: A piece to retrieve relevant documents for the question posed, known as the Document Retriever, and the Document Reader, a multi-layer recurrent neural network that extracts the answer span from the retrieved documents. The Document Reader's input are individual tokens (represented as 300-dimensional vectors), along with their part-of-speech tag and named entity type (encoded as inputs of 50 and 19 elements respectively).

This paper describes experiments carried out to evaluate the use of HRR to compress each token, its part-of-speech, and its named entity type (if any), into a single 300-dimensional vector. Compared to DrQA, this decreases the width of the input data from 369 to 300, a reduction of over 18%, thereby speeding up training. In Section 2, the background and prior related work is described. Section 3 presents the holographic reduction process, including the slot and filler scheme employed, and describes the modifications made to the Document Reader portion of DrQA, to enable it to process the holographically compressed tokens. The framing of the experiments are found in Section 4, with results and analysis are given in Section 5. The conclusions and future research avenues are summarized in Section 6.

## 2. BACKGROUND AND RELATED WORK

This research is based primarily on two prior works: holographic reduced representations, and the DrQA question answering architecture. Each of these is described in turn below.

## 2.1. Holographic Reduced Representations

Holographic reduced representations (HRR) apply algorithms from the field of signal processing, namely circular convolution and its approximate inverse, circular correlation, to compress multiple vectors of dimension *n*, into a single vector having that dimension [3]. In order to accomplish this, elements of the initial vectors are superposed in the resulting vector. While superposition can also be achieved by averaging two vectors element-wise, the average operation is not invertible. Circular convolution, on the other hand, also results in a superposed vector, but is an invertible operation. Circular convolution applied to vectors **c** (having elements $c_0, …, c_n$) and **x** (with elements $x_0, …, x_n$), yields a vector **t**, having elements $t_0, …, t_n$. Figure 1 shows the circular convolution compression of two vectors (*c* and *x*), each having three dimensions, into a single three-dimensional vector, *t*, whose elements are computed as depicted.

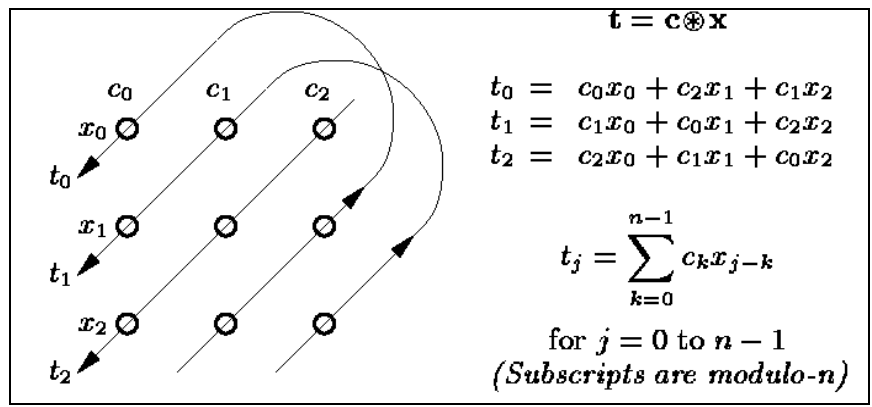

Figure 1. Circular convolution (adapted from [3]).

Multiple holographically compressed vectors may be composed together by element-wise summing or averaging. Thus, the composition operation enables compression of more than two vectors into a single vector having the same dimension as a single constituent component.

Circular convolution is a commutative operation, and as such cannot be used to represent complex structures, such as sequences or hierarchies, since order is lost. A recommended approach for representing a complex structure such as a frame with slots, is to create a frame label vector, and sum it with one or more slot label vectors, circularly convolved with their respective filler vectors [3]. Various aspects of working with HRR vectors require that they be (nearly) orthogonal. Pre-trained word vector embeddings do not necessarily have this orthogonality property.

This research effort makes use of a holographically reduced frame and slot structure, to represent the token, its part-of-speech tag, and named entity type. The approach followed results in quasi-orthogonal vectors, and is fully described in Section 3.

## 2.2. DrQA

The DrQA question answering framework developed by Facebook Research [8] was designed to answer open domain questions from large sources of information like Wikipedia. It is divided into two primary components: the Document Retriever retrieves documents (from Wikipedia or other sources) that may contain the answer to the question; the Document Reader identifies the answer span from the retrieved text documents. In the original work, this latter component was evaluated using SQuAD 1.1, a dataset consisting of a number of passages and over 100,000 questions, whose answers are extractable from the passages as spans of text [9]. The experiment described in this paper only makes use of the Document Reader and the SQuAD dataset. The Document Retriever is not considered.

The neural network used in the Document Reader is a composite of multiple neural networks, each having a distinct function. Tokens in both the question and the passage are encoded and presented to the network. The **question** encoding simply takes the 300-dimensional *GloVe* embedding for each token, passes it through a bidirectional LSTM, and combines the vector from the hidden units via a weighted sum (whose weights are learned by the network). The **passage** encoding is more complex, with each token represented by:

- Its 300-dimension *GloVe* vector
- Its part-of-speech tag
- Its named entity type (if any)
- A 4-element feature set, indicating: 1) whether the token matches a token in the question exactly; 2) whether there is a case-insensitive match between the token and a token in the question; 3) whether there is lemma form match between the token and a token in the question; and 4) the token's term frequency within the passage.

The above described representation for each passage token, and an attention-based real-valued weighted measure (derived from a simple feed-forward network) which provides the similarity between related but morphologically dissimilar words in the passage and question, such as *hit* and *single* in the music (or baseball) sense, is run through a bidirectional LSTM. That output, along with the weighted sum question representation, is fed to two bilinear attention classifiers: one predicts the probability of the passage token being the start of the answer span, and the other predicts the probability of it being the end of the answer span. The network's output is a pair: the passage token positions having the highest probabilities for being the start and end of the answer span. The architecture of the Document Reader neural network is shown in Figure 2, and depicts the partial representation of the question *When did Beyoncé release Dangerously in Love?* and the portion of the passage containing the answer, that reads in part … *Beyoncé's debut album, Dangerously in Love, … .*

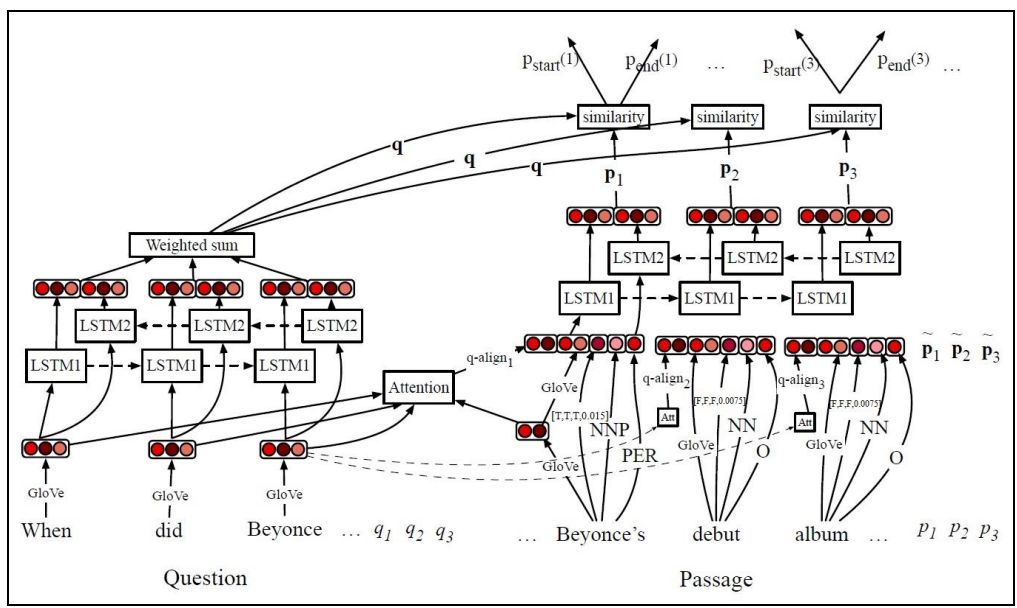

Figure 2. DrQA Architecture (adapted from [10]).

### 2.3. Related Work

A search of literature yielded no instances of works that use HRR with DrQA, thus establishing the novel aspect of this research. Only one prior work in general question answering utilizing HRRs with neural networks was found [11]. In those experiments, the goal was to detect the relationship between questions and answers (no mention of passages or context documents is made). Their architecture differs significantly from this research, first passing pre-trained embeddings for questions and answers into separate LSTMs, and passing the last hidden state from each to a layer where a holographic representation is constructed for further processing. An additional dissimilarity from this work is that circular correlation was used in constructing the holographic representation, rather than circular convolution.

The Document Reader architecture was modified for this research as shown in Figure 3. The passage input was changed to remove the previously separate encodings of parts-of-speech and named entity types. These attributes are now compressed into the token embeddings. The question input stage was altered to accept the compressed tokens which include part-of-speech and named entity type, unlike in the original architecture. All other aspects were retained.

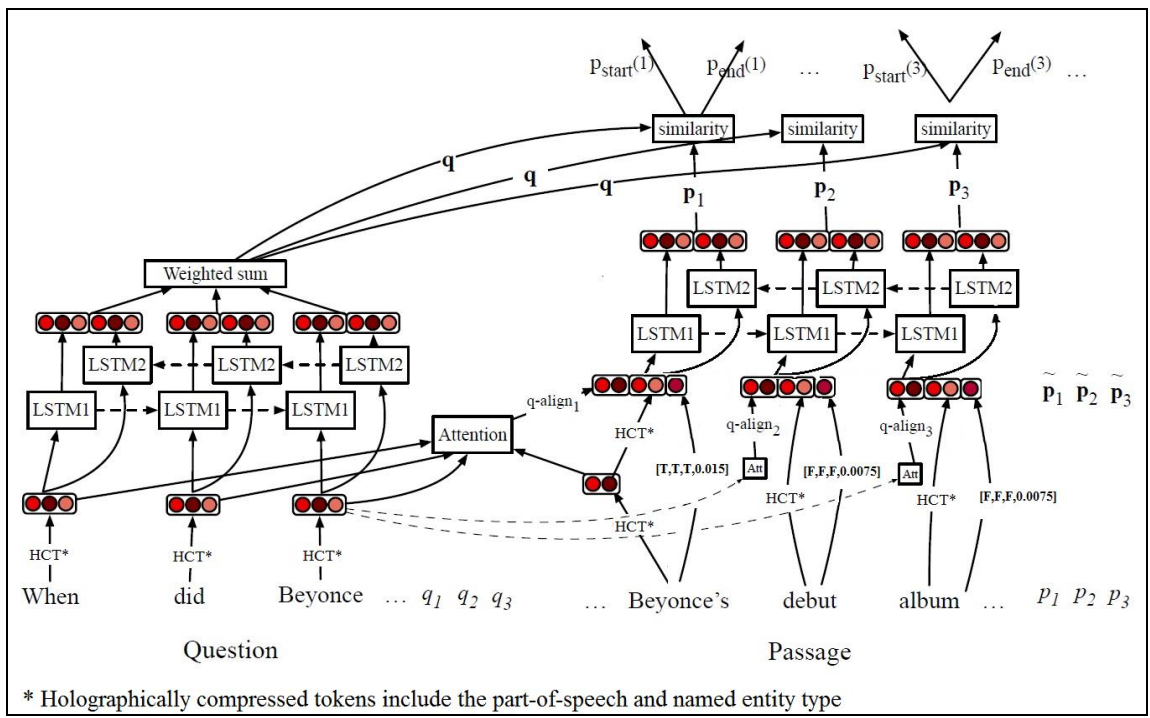

Figure 3. Modified Architecture.

## 3. METHODOLOGY

The experiments crafted for this research were designed to answer specific research questions:

Research question #1 – Can pre-trained word vector embeddings (which are not guaranteed to be orthogonal to each other) be made (nearly) orthogonal by constructing holographic frame and slot structures with them? The answer to this question is important, as the theory of holographic reduced representations relies on vectors being orthogonal to each other.

Research question #2 – Do holographically compressed token embeddings retain the semantic relationships they possessed pre-compression? The value of embeddings pre-trained on large corpora comes from the semantics that are learned from the various contexts in which words appears. To take advantage of this benefit, those relationships must generally be preserved.

Research question #3 – How do compressed tokens (consisting of a holographic frame-slots structure for the token, its part-of-speech, and its named entity type) perform, compared to uncompressed embeddings with separate features for part-of-speech and name entity type, in the DrQA Document Reader architecture? To hold promise for future applications, the accuracy metrics of the proposed compression approach must be close or superior to those of the unmodified method.

The methodology required work along two fronts. First, the pre-trained *GloVe* embeddings had to be converted into their new holographically compressed embeddings form. After this was done, the Document Reader architecture was adapted to accept the new input format. The approach followed in each of these phases is detailed in the subsequent two sections.

### 3.1. Holographically Compressed Tokens

The method undertaken to compress a word token was to create a frame and slots structure, as recommended in the seminal work [3]. In order to do this, several 300-dimensional vectors, were randomly generated to represent labels for the frame and each slot, as well as each part-of-speech tag and each named entity type. The structure was then constructed from the pre-trained *GloVe* 300-dimensional token vectors, and vectors for the frame label (HRRTOK), the token slot (TOK), the token's part-of-speech tag slot (POS), and the token's named-entity type slot (ENT), if one exists. The holographically compressed token ($TOK_{HC}$) in this work is calculated as:

$$TOK_{HC} = ( HRRTOK + TOK \otimes token + POS \otimes pos\_tag + ENT \otimes ne\_type ) / m$$

where *token*, *pos_tag*, and *ne_type* are vectors for the fillers: the token, its part-of-speech, and named entity type (if applicable), and $\otimes$ is the circular convolution operator. The value *m* is the number of vectors summed (in the above example four vectors are summed). The division by this factor normalizes the compressed vectors. Tokens that do not have a named entity type are normalized by dividing by three). As an example, the token *IBM* would be represented as:

$$IBM_{HC} = ( HRRTOK + TOK \otimes \mathbf{\mathit{IBM}} + POS \otimes \mathbf{\mathit{NNP}} + ENT \otimes \mathbf{\mathit{ORG}} ) / 4$$

The italicized and bolded fillers above are the vectors for the token (IBM), its part-of-speech (NNP is the tag for proper noun), and named-entity type (ORG = organization). The vectors for the frame and slot labels, each part-of-speech tag, and each named entity type were randomly selected from a normal distribution with mean 0 and variance $1/n$, where *n* is the dimensionality of the vectors (300 in this case). While reduced holographic representations are generally presented with vectors having dimensionality in the thousands, in this study the randomly drawn vectors with 300 dimensions were sufficiently orthogonal, and their composition with the pre-trained *GloVe* embeddings yielded a vocabulary represented by vectors nearly orthogonal to each other, with the angle separating most pairs of vectors being greater than 75 degrees.

### 3.2. DrQA Document Reader Modifications

The source code for DrQA is available at https://github.com/facebookresearch/DrQA, under a BSD license. Since only the Document Reader was to be used, a version modified to run that

Computer Science & Information Technology (CS & IT)                                 245

component stand-alone was downloaded from https://github.com/hitvoice/DrQA. This baseline uses the spaCy natural language processing framework (https://spacy.io/) to determine parts-of-speech and named entity types (the original uses the Stanford CoreNLP suite). *GloVe* embeddings of 300 dimensions, trained on the 840 billion word corpus, were used.

The source code was downloaded and modified to accommodate the holographically compressed token format. Changes were made to:

- The preprocessing stage (this is where grammatical annotations are made and the vocabulary of the SQuAD dataset is mapped to embeddings) – Since the compressed tokens include the part-of-speech tag and named entity type, this meant that a simple word type might no longer be distinct, as those attributes might be dissimilar in different contexts. The raw token was constructed from the word type (in lowercase), concatenated with its part-of-speech and named-entity type. As an example, the word type *fish* would be converted in the sentences below to the word type shown in parentheses:

    *I like to fish early in the morning. (fishVB) - fish is a verb*
    *That was the best fried fish I have ever eaten. (fishNN) – fish is a noun*
    *This is Dr. Fish. (fishNNPPERSON ) – fish is a proper noun with type PERSON*

    While *fish* has a single pre-trained *GloVe* embedding, it maps to distinct compressed vectors in the above examples. Additionally, no tokens are left without representation in this research, as tokens that were not in *GloVe* were mapped to a special vector for unknown word types. These two facts lead to an increase in the size of the vocabulary of the holographically compressed tokens. In the experiment, the size of the vocabulary grew from 91,590 to 158,454 distinct tokens. Each modified token was used as an index that maps it to its holographically compressed vector, which is fed to the neural network.

    The other change made to the preprocessing stage was the addition of holographic compression for question tokens. It should be noted that this differs from the baseline model, which does not use part-of-speech and named entity types for question tokens. However, this change was made to keep the representations coherent for the attention mechanism in the network.

    The 4-tuple feature set of binary similarities and term frequency was retained for processing without modification.

- The neural network – Changes were made to the neural network to remove the paths for input of the explicit part-of-speech tag and named-entity type features associated with passage tokens, and to set it up to process only the compressed tokens and the 4-tuple. No changes were required to the paths that process the question tokens, as those simply accept a 300-dimensional vector. The difference in this case is that the holographically compressed vectors for the question tokens also contained the part-of-speech tag and named-entity type.

## 4. EXPERIMENTS

The experiments were constructed to facilitate a valid comparison between the DrQA Document Reader in its original form (using only the pre-trained *GloVe* embeddings) and the holographically compressed tokens constructed in this research. This included following the model validation approach undertaken for DrQA.

**4.1. SQuAD Input Data**

The SQuAD 1.1 question answering dataset [9] was used for this research. In this dataset, each answer is a span from the passage (or context) from which the question is taken. The training set used was made up of over 87K questions, and the validation data had over 10K questions.

**4.2. Experimental Setup**

The experiments were carried out in the same manner as was done for DrQA, including the validation approach. The experiments were run twice: once for the baseline (unmodified) Document Reader, and once using the holographically compressed tokens. Each experiment ran for 40 epochs.

**4.3. Model evaluation and data collection**

An evaluation of the model was completed using the validation set, at the end of each epoch, as was done in the original work. For consistency, two accuracy measures were collected at the completion of all epochs: an exact match metric (the predicted span was correctly identified with both the start and end indices), and a broader F1 measure for when the answer was correct, but incomplete (e.g., the correct answer spanned the string *John Smith*, but the model selected the span for either *John* or *Smith*).

## 5. RESULTS AND ANALYSIS

This section contains the results and analysis of the experiments. The findings are presented as they relate to each of the three research questions, in sections 5.1 through 5.3 respectively.

**5.1. Orthogonality of the Holographic Vocabulary Vectors**

In order to compose complex structures, such as sequences that represent a sentence, the resulting holographically compressed tokens, must be (nearly) orthogonal. While the experiments done here do not represent phrases or sentences, this effort was undertaken as a step toward that capability. Given the size of the final vocabulary, a full pairwise set of distances between the vectors was too computationally expensive to assess. Instead, to evaluate the orthogonality of the embeddings, two disjoint lists of 100,000 tokens were selected at random, and the cosine similarity for tokens at the same index position in those lists was calculated. In nearly 94% of these cases, the cosine similarity was less than 0.25 (less than 15 degrees from fully orthogonal). Since orthogonality is not an attribute of the *GloVe* embeddings, this characteristic is directly attributable to the holographic compression attained through the orthogonality of the frame and slot labels vectors used.

**5.2. Preservation of Semantic Relationships**

As was previously mentioned, the benefit of embeddings trained on large corpora is the robust set of semantic relationships extracted from word usage in context. It would be desirable to preserve these semantic relationships to the greatest extent possible. In the experiments, vectors were mapped to different vector bases by the holographic reduction process, making it impossible to retain the exact relationships that existed between tokens prior to their transformation. The approach taken in evaluating the preservation of semantic relationships was to select core tokens randomly, and to analyze the composition of their neighborhoods (which tokens are closest it?) in both bases. In each of the next figures, the core token is encircled and its ten closest neighbors in both spaces are shown. The data is visualized in two dimensions by use of the t-SNE technique [12].

Figure 4, depicting the token *radially*, shows a case where the semantic positions are largely preserved, and only one token differs between the methods. The holographically reduced neighborhood does not include the embedding for *actuating*, adding in its stead a different, but semantically related, token (the embedding for *protrusion*).

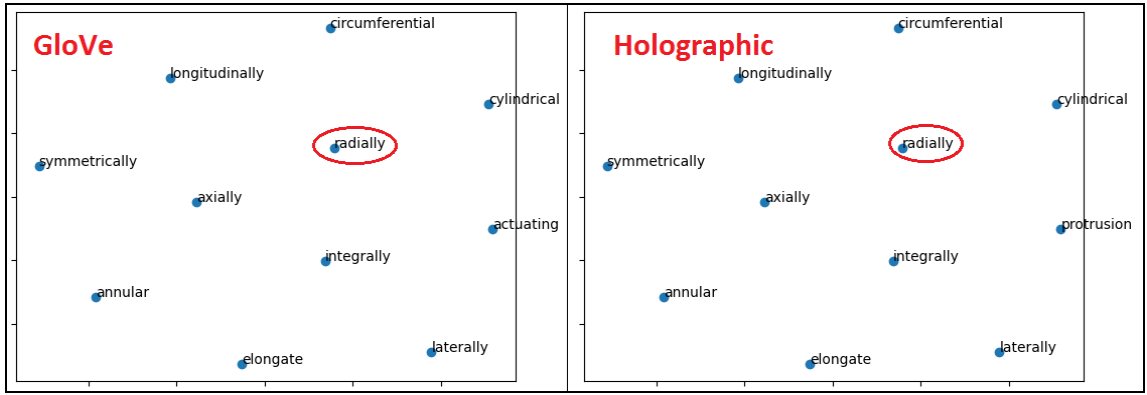

Figure 4. Vector neighborhoods for the token *radially*.

The neighborhood for the token *sprint* is shown in Figure 5. Most neighbors are similar in the two cases, with tokens *Sprint* and *sprints* holding the same relative positions. However, the relationships of many neighbors shifted (*javelin*, *moto*, *race*, *racing*, *sprinter*, and *sprinting*). The holographic embeddings omit tokens *triathlon* and *racer*, and include *champion* and *relay*.

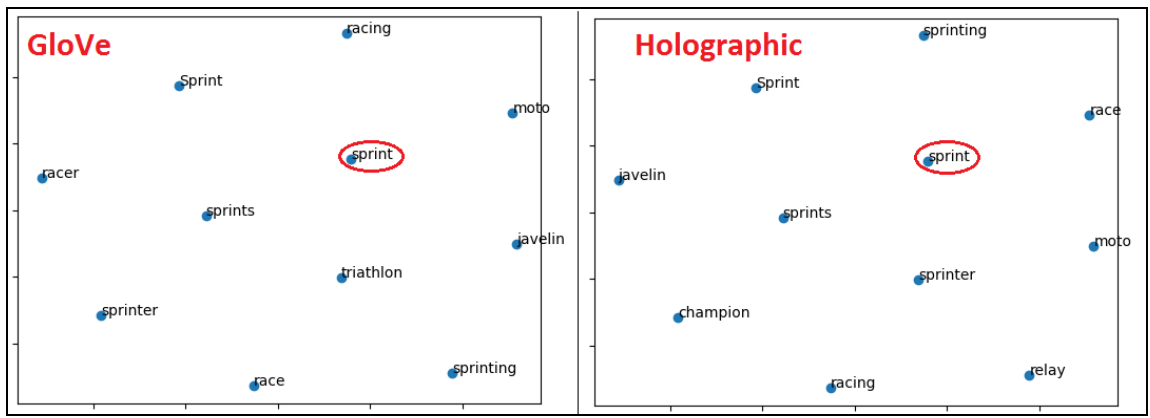

Figure 5. Vector neighborhoods for the token *sprint*.

Finally, Figure 6 portrays the neighborhoods for token *shared*. This was a case where there were significant differences between the two sets of embeddings. Only four neighbors are in common, with two holding their relative positions (*share* and *sharing*), and two showing some displacement (*own* and *Shared*). Tokens *both*, *communal*, *experiences*, *family*, *personal*, and *together* do not appear in the holographic neighborhood, which instead includes embeddings for *access*, *contributed*, *ideas*, *linked*, *others*, and *realized*.

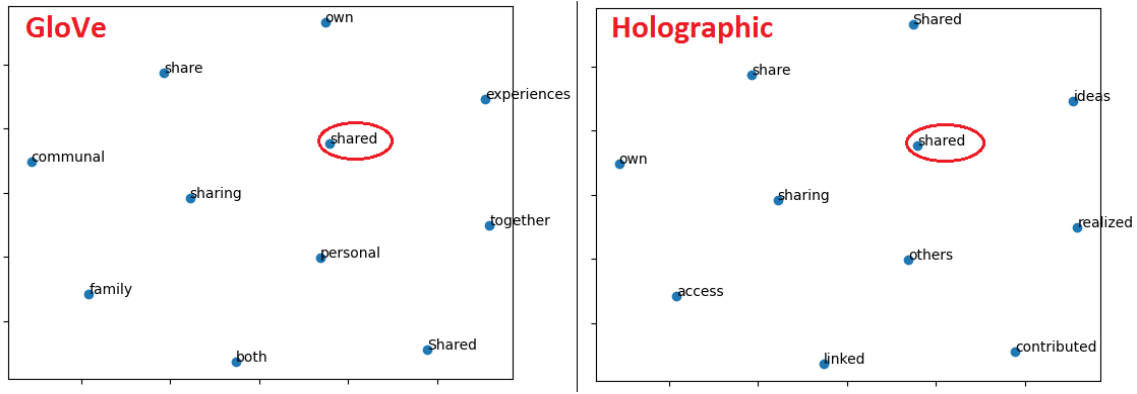

Figure 6. Vector neighborhoods for the token *shared*.

A detailed pairwise analysis was done on the composition of the neighborhoods. Specifically, the ten neighbors were sorted by their distance from the core token in both the unmodified and holographically compressed cases. The results showed that approximately: 18% of the vectors in each neighborhood were in the same relative position in both cases; 43% were present in both neighborhoods, but were shifted in distance; and 39% of words in the neighborhoods were disjoint between the two sets. In the disjoint cases, the departures seen were deemed reasonable by a manual review. Although the words that differ are semantically related to the core token, additional research is needed to better characterize these differences and their potential impact on applications.

### 5.3. Accuracy Metrics

Training took place over 40 epochs for each case, and the best model was selected based on the F1 score. For the unmodified Document Reader, F1 was 78.23%. That model yielded an exact match score of 68.83%. These values are consistent with those reported in the original DrQA research [8]. The holographically compressed approach's best model had an F1 measure of 76.28%, with exact match at 66.70%. The scores are summarized in Table 1.

Table 1. Summary of Accuracy Metrics.

|  | F1 | Exact Match |
| --- | --- | --- |
| Unmodified Document Reader | 78.23% | 68.83% |
| Holographically Compressed Tokens | 76.28% | 66.70% |

While the performance attained fell short of the original work, it did not do so by much: both the exact match measure and the F1 metric were around two percentage points below the level for DrQA Document Reader. Furthermore, this research effort mirrored the approximately 10% difference seen between these two measures in the original case. As an initial effort without detailed optimization, the results point to a promising method that has the potential to be refined for better results.

On the positive side, the time for completion of individual epochs decreased by over 30% in the holographically compressed tokens case, due to the reduction in the input data width. Notwithstanding the added 22 minutes of preprocessing time (a one-time cost) incurred as a result of compressed token construction and a larger vocabulary, the faster training resulted in a savings of more than 75 minutes over the 40 epochs. This increased performance in speed is significant since it allows for more training epochs to be completed in the same period of time.

## 6. CONCLUSIONS AND FUTURE WORK

The experiments in this research explored the viability of holographic reduced representations, as a means for compressing additional information into a token's fixed-dimensional vector embedding. To relate this effort to prior work for which a comparison baseline existed, an existing question answering framework was modified to evaluate the proposed methodology. The experiments done provide a launching point for future research in other deep learning frameworks, and for other natural language processing tasks.

### 6.1. Limitations

The expansion of the vocabulary size is the main limitation of the proposed approach. As an example, if additional information, like the output of a dependency parser, was to be holographically compressed with the token, its part-of-speech, and named-entity type, the vocabulary could become prohibitively large. This could prevent referring to stored embeddings with an index, a common practice in natural language work. Although this constraint could be circumvented by feeding the computed embeddings directly to the neural network, without first storing them in an embedding layer (that could be loaded as needed), this is undesirable for large problems as it would necessitate completion of both preprocessing and training/evaluation in single runs of the model.

### 6.2. Conclusions

The main contributions of this work fall in three areas: 1) construction, using HRR, of nearly orthogonal vectors representing a token's pre-trained embedding, its part-of-speech, and its named entity type, 2) documentation of results of experiments assessing the preservation of semantics after applying HRR, and 3) the favorable performance metrics attained, when compared to the baseline system. The fact than this non-optimized holographic compression model came within two percentage points of the original Document Reader, makes this a promising method for further investigation and improvements. Although one might have expected that the expansion of the size of the vocabulary by over 70% would degrade the accuracy significantly, this was not the case.

When combined with an over 30% speedup in training time, these experiments clearly demonstrate that holographically compressed token embeddings are viable in deep learning for question answering, and warrant further investigation in this and other application areas.

### 6.3. Future Work

This work was undertaken to set the stage for further investigation into holographically compressed token embeddings, by evaluating the necessary condition of the orthogonality of its representations, and measuring the preservation of pre-trained semantic relationships. Possible avenues for future work include:

- Characterize the semantic shifting that happens as a result of using HRR. Investigate their direct impact and identify potential remedies.
- Study the impact on performance, of compressing additional information into the token's representation. Information from a dependency parse is a good candidate.
- Remove the 4-tuple feature set from the input data. In the original work, removal of these features had very little impact on the metrics.

- Devise a method to encode sequences, such as sentences, with HRR, and use those representations as input to the network (rather than word tokens). This exploration would likely be need to be accompanied transitioning to a different (non-RNN) neural network architecture.
- Assess compressed tokens in different NLP applications. The proposed methodology was applied to question answering due to the availability of a baseline that made use of embeddings semantics and of elements that could be compressed into a single vector (part-of-speech tags and named entity typed). However, it lends itself to exploration in many other areas.

## ACKNOWLEDGEMENTS

This paper was supported by Faculty Research & Creative Activity Committee (FRCAC) grant number 221751 at Middle Tennessee State University. The author is grateful for this support.

## REFERENCES

[1] Mikolov, T., Sutskever, I., Chen, K.,Corrado, G. S., and Dean, J. (2013). "Distributed representations of words and phrases and their compositionality", In NIPS 13, pp. 3111–3119.
[2] Pennington, J., Socher, R., & Manning, C. D. (2014). "Glove: Global vectors for word representation". In Proceedings of the 2014 Conference on empirical methods in natural language processing (EMNLP), pp. 1532-1543.
[3] Plate, T.A., (1995). "Holographic reduced representations". IEEE Transactions on Neural networks, 6(3), pp.623-641.
[4] Hochreiter, S. & Schmidhuber, J.,(1997). "Long short-term memory". Neural computation, 9(8), pp.1735-1780.
[5] Lee, K., Salant, S., Kwiatkowski, T., Parikh, A., Das, D. and Berant, J., (2016). "Learning recurrent span representations for extractive question answering." arXiv preprint arXiv:1611.01436.
[6] Wang, W., Yan, M. and Wu, C., (2018). "Multi-granularity hierarchical attention fusion networks for reading comprehension and question answering". arXiv preprint arXiv:1811.11934.
[7] Nakov, P., Hoogeveen, D., Màrquez, L., Moschitti, A., Mubarak, H., Baldwin, T. and Verspoor, K., (2019). "SemEval-2017 task 3: Community question answering". arXiv preprint arXiv:1912.00730.
[8] Chen, D., Fisch, A., Weston, J., & Bordes, A. (2017). "Reading wikipedia to answer open-domain questions". arXiv preprint arXiv:1704.00051.
[9] Rajpurkar, P., Zhang, J., Lopyrev, K. and Liang, P., (2016). "SQuAD: 100,000+ questions for machine comprehension of text". arXiv preprint arXiv:1606.05250.
[10] Jurafsky, D. & Martin, J. (2019) *Speech and Language Processing*, 3$^{rd}$ Edition Draft, dated October 16, 2019. Available online at https://web.stanford.edu/~jurafsky/slp3/ . Accessed on June 11, 2020.
[11] Tay, Y., Phan, M.C., Tuan, L.A. and Hui, S.C., (2017). "Learning to rank question answer pairs with holographic dual lstm architecture". In Proceedings of the 40th international ACM SIGIR conference on research and development in information retrieval, pp. 695-704.
[12] Maaten, Laurens van der, and Geoffrey Hinton. (2008). "Visualizing data using t-SNE." Journal of machine learning research 9 (Nov), pp. 2579-2605.